*Original Article*

# Novel Regression and Least Square Support Vector Machine Learning Technique for Air Pollution Forecasting

M. Dhanalakshmi[1], V. Radha[2]

[1,2]Department of Computer Science, Avinashilingam Institute of Home Science and Higher Education for Women, Coimbatore, India.

[1]Corresponding Author : dhanalaxmi2289@gmail.com

***Abstract*** *- Air pollution is the origination of particulate matter, chemicals, or biological substances that brings pain to either humans or other living creatures or instigates discomfort to the natural habitat and the airspace. Hence, air pollution remains one of the paramount environmental issues as far as metropolitan cities are concerned. Several air pollution benchmarks are even said to have a negative influence on human health. Also, improper detection of air pollution benchmarks results in severe complications for humans and living creatures. To address this aspect, a novel technique called, Discretized Regression and Least Square Support Vector (DR-LSSV) based air pollution forecasting is proposed. The results indicate that the proposed DR-LSSV Technique can efficiently enhance air pollution forecasting performance and outperforms the conventional machine learning methods in terms of air pollution forecasting accuracy, air pollution forecasting time, and false positive rate.*

## 1. Introduction

Air pollution presents a significant health problem as far as urban metropolises are concerned. Though air pollution monitoring and forecasting accurately and precisely are found to be tremendously critical, prevailing data-driven methods have so far entirely acquired the complicated interactions between spatial and temporal aspects of air pollution. Moreover, uneasiness for the environment, health, and welfare has fascinated substantial global awareness owing to the new environmental confronting that menace the planet.

A method called Deep-AIR employing CNN and LSTM framework was proposed by Qi Zhang et al. [1] to fill the gap in ensuring fine-grained city-wide air pollution estimation by means of domain-specific features that, in turn, captured spatio-temporal features. Also, a 1 to 1 convolution layer was designed with the purpose of improving the learning of temporal and spatial interaction. As a result, the forecasting accuracy for air pollution was improved.

Variational Auto Encoder (VAE) based on the innovative Integrated Multiple Direct Attention Deep Learning architecture (IMDA) [VAE-IMDA] was proposed by Abdelkader et al. [2], taking into consideration the conventional VAE and attention mechanism to forecast distinct air pollutants in a computationally efficient and accurate manner. Also, the temporal dependencies between nonlinear approximations concentrating on the relevant feature extraction were ensured.

To this extent, this paper is proposed for a research study on applying Discretized Regression and Least Square Support Vector (DR-LSSV) based air pollution monitoring and control for IoT networks. In this study, we proposed a system for predicting air quality by advanced machine learning model - hybrid Discretized Regression and Least Square Support Vector air pollution forecasting. Two baseline models were also built for comparison with our proposed method.

## 2. Literature Review

Up till now, it remains a major issue in acquiring optimal and accurate air pollution estimation with high accuracy and low false positive rate simultaneously. Numerous missing values are said to exist in both temporal and spatial dimensions. This, in turn, has resulted in severely undermining the performance of optimization-based air pollution methods.

A two-layer model prediction method based on Long Short-Term Memory Neural Network and Gated Recurrent

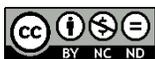

Unit called LSTM&GRU was proposed by Baowei et al. [3]. A double-layer Recurrent Neural Network method was designed with the objective of predicting the PM2.5 value. With this two-layer model, a better prediction rate was achieved. A novel deep learning-based air quality forecasting by learning spatial-temporal correlation features and multivariate air quality-related interdependence time series data employing hybrid deep learning architecture was proposed by shengdong et al. [4]. Satisfied accuracy was said to be achieved via a hybrid learning model. Despite accuracy being attained, the precision was not focused. To concentrate on this issue, a hybrid CNN-LSTM model was proposed by Shreya K et al.[5] that not only ensured accurate prediction via extraction of spatiotemporal features using CNN but also improved the predictive performance via LSTM-based deep learning.

The ceaseless precipitation of global urbanization and industrialization has led the way towards several issues. One of the important environmental issues is air quality, persuaded by the deployment of urbanization and industrialization. In Yuting et al.[7], explainable deep learning model was introduced with the objective of improving the prediction accuracy of air quality. In Qing et al.[8], yet another short-term forecasting model employing convolutional-based bidirectional gated recurrent unit (CBGRU) was proposed therefore minimizing the error involved during forecasting. A novel method to analyze deep air quality by utilizing convolutional neural and long short-term memory was proposed by Ekta et al.[9].

The prevailing and recent methods, however, only concentrated on giving detail about their causes and temporal associations. However, a holistic view of the pure forecasting performances was not analyzed. In Raquel et al. [11], a methodology was designed based on two criteria, namely, exactness and robustness, to make an elaborate comparison of distinct pollutant forecasting methods and their characteristics. With this high influence on accuracy was said to be arrived at.

Among numerous air pollutants, Particulate Matter of diameter less than 2.5μm is one of the most serious health issues. It results in different types of illnesses in the respiratory tract and cardiovascular diseases. Therefore, it becomes highly necessary to predict PM2.5 concentrations accurately so that citizens can be circumvented the dangerous influence of air pollution in the initial stage. In abdellatif et al. [12], a deep learning solution was designed with the purpose of predicting the PM2.5 hourly forecast in Beijing, China. This work designed a hybrid model combining CNN-LSTM where spatial-temporal features were extracted by integrated historical pollutants, specifically, PM2.5 concentration in the neighboring stations. A case study of air pollution prediction using machine learning was investigated in detail by kumar et al. [13].

In Azim et al. [15], a novel new hybrid intelligent method was designed based on long short-term memory (LSTM) and a multi-verse optimization algorithm (MVO). With this hybrid intelligent method, air pollution prediction and analysis were made by utilizing the data from Combined Cycle Power Plants (CCPP). Moreover, a long short-term memory model was also utilized with the purpose of forecasting the amount of NO2 and SO2 by CCPP. Also, the MVO algorithm was employed for optimizing LSTM parameters to minimize errors involved in forecasting.

With several features involved in predicting air quality data, a small deviation would cause a greater amount of error. To address this aspect, a method for predicting air quality based on multiple data features by fusing multiple machine learning models was presented by Ying et al. [16]. With this, not only accuracy was ensured but also minimized the loss involved during air pollution forecasting. In Yue-shan et al. [17], an Aggregated LSTM model (ALSTM) was proposed by integrating local air quality monitoring stations, i.e., the station in adjacent areas and obtained from external pollution sources were combined. Also, to enhance the prediction accuracy, three LSTM models were aggregated for early predictions based on external sources of pollution obtained from adjacent industrial air quality stations. As a result, the prediction accuracy was also improved to a greater extent.

Owing to the vigorous, dynamic phenomenon and several spatio-temporal factors influencing air pollution dispersion and involving uncertainty estimates makes, it is trustworthy. This, in turn, assists the decision-makers in taking proper actions concerning the pollution crisis. In Ichrak et al. [19], a multi-point deep learning method was proposed by considering convolutional long short-term memory (ConvLSTM) for the large arbitrary nature of air quality forecasting. As a result, even accuracy was arrived at even though with the huge climatic change.

As far as the meteorological forecast and air controlling are concerned, air quality prediction is contemplated as the paramount reference. However, with the occurrence of overfitting in prediction algorithms based on a single model, complexity is also said to be increased. To address this aspect, a prediction method based on integrated dual LSTM was proposed by Hongqian et al.[20]. With this, the precision of prediction data was said to be improved greatly. A recurrent neural network with LSTM was integrated by Saba et al. [21] with the purpose of not only reducing the error but also improving forecasting accuracy to a greater extent.

However, another method to minimize overfitting employing an ensemble network (EN) that combines recurrent neural network (RNN), LSTM network and gated recurrent unit (GRU) network was designed by Canyang et al. [22]. With this ensemble type of networking, accuracy with the mean absolute error was improved greatly. In Jun et al. [23], urban ecological monitoring focused on deep learning to improve air pollution forecasting.[14] To identify nonlinear relationships between input and output variables, wavelet neural networks were integrated with meteorological conditions for air pollution forecasting by Qingchun et al. [24]. To address air pollution monitoring and control via machine learning technique[6], a Linear Regression and Multiclass Support Vector (LR-MSV) IoT-based Air Pollution Forecast method provide forecasts in an accurate and timely manner with minimum error rate was designed in Dhanalakshmi et al. [12].

In this paper, by a comparison of conventional machine learning and classical deep learning models, a novel air quality forecasting method, called Discretized Regression and Least Square Support Vector (DR-LSSV), is proposed. It is motivated to address error minimization and improve accuracy by utilizing the Discretized Hartley Transformation and performing feature selection using Constrained Maximum Likelihood Linear Regression. Finally, the proposed DR-LSSV method can classify air quality-related time series data by means of the air quality index and the selected features via the Concordance Correlative function for different weather conditions and different traffic states both on an hourly and daily basis.

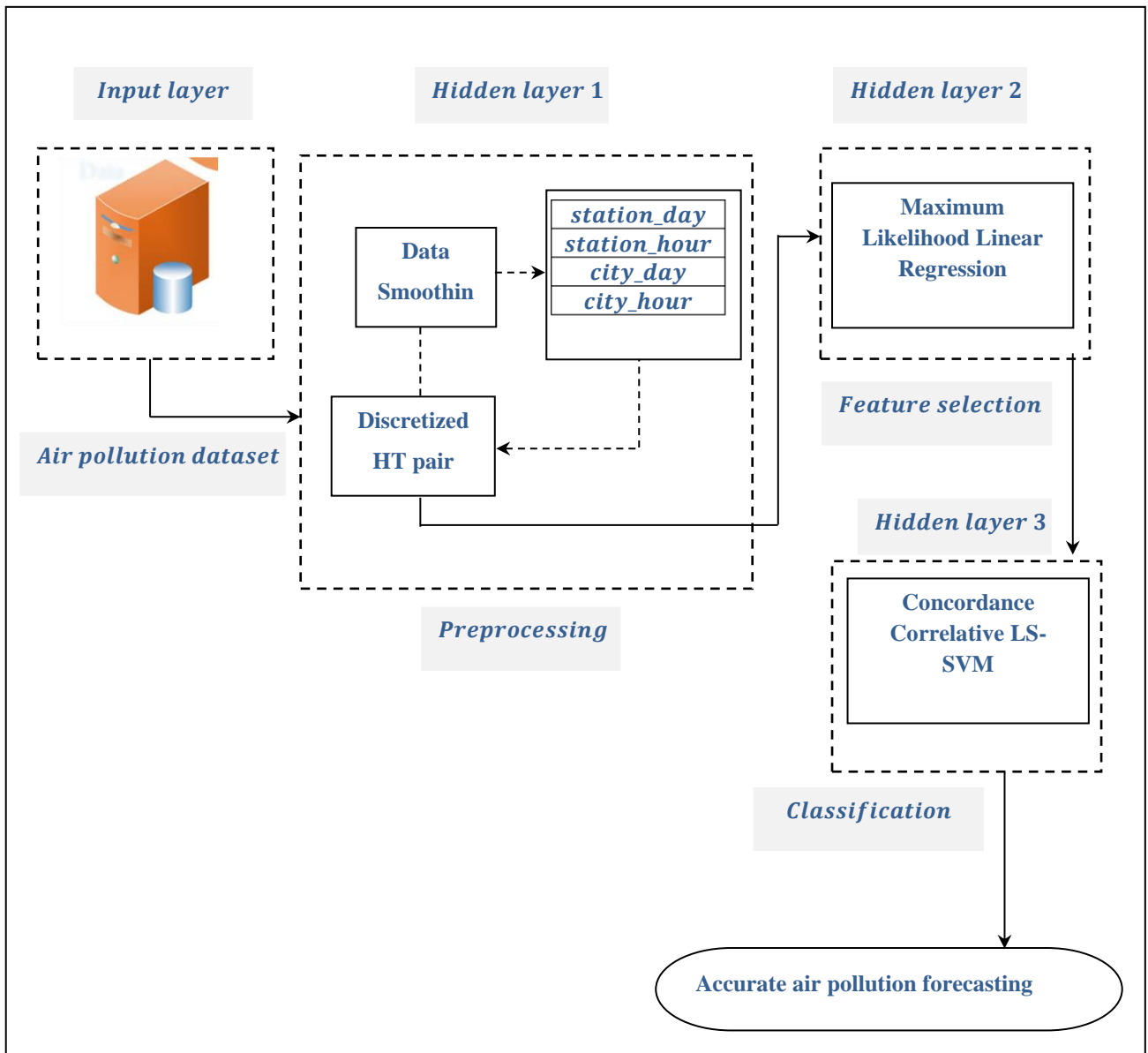

**Fig. 1 Architecture of Discretized Regression and Least Square Support Vector**

## 3. Methodology

Air pollution forecasting refers to the application of science and technology with the objective of predicting the air pollution composition present in the atmosphere for a specific location and time. Over the past decades, air pollution has been considered the world's considerable issue. It brings about respiratory-related issues, cardiovascular concerns, lung diseases, and therefore resulting in mental-related affairs and provoking prevailing health state of affairs. Hence, minimizing the making of people well informed of these issues emanated by air pollution becomes indispensable. In this work, a novel method called Discretized Regression and Least Square Support Vector (DR-LSSV) is proposed in this work to increase air pollution forecasting accuracy with minimizing error. Figure 1, given below, shows the overall architecture of the DR-LSSV method.

As shown in the above figure, the proposed DR-LSSV method consists of three different processes, namely preprocessing, feature selection, and classification, with multiple layers involved, such as one input layer, three hidden layers, and one output layer. Air quality data in our work are obtained by means of IoT devices. The collected air quality data are provided as input to the input layer. Air quality data obtained as input is provided in the first hidden layer, where preprocessing is carried out with the aid of Discretized Hartley Transformation function to remove the noise present in the raw air quality data.

Following the second hidden layer, the feature selection process is carried out using the Constrained Maximum Likelihood Linear Regression function that, in turn, selects significant features. Finally, the classification is performed in the third hidden layer by applying Concordance Correlative Least Square Support Vector to obtain final classification results by analyzing testing and training data. Based on the obtained results in the output layer, accurate air pollution forecasting is achieved with minimum error.

### 3.1. Data Collection

This research work gathers information from Air Quality Data in India for the period between 2015 and 2020 that consists of Air Quality Index (AQI) at the hourly and daily levels of numerous stations in multiple cities in India. The AQI calculation utilizes 7 measures, namely, PM2.5, PM10, SO2, NOx, NH3, CO and O3.

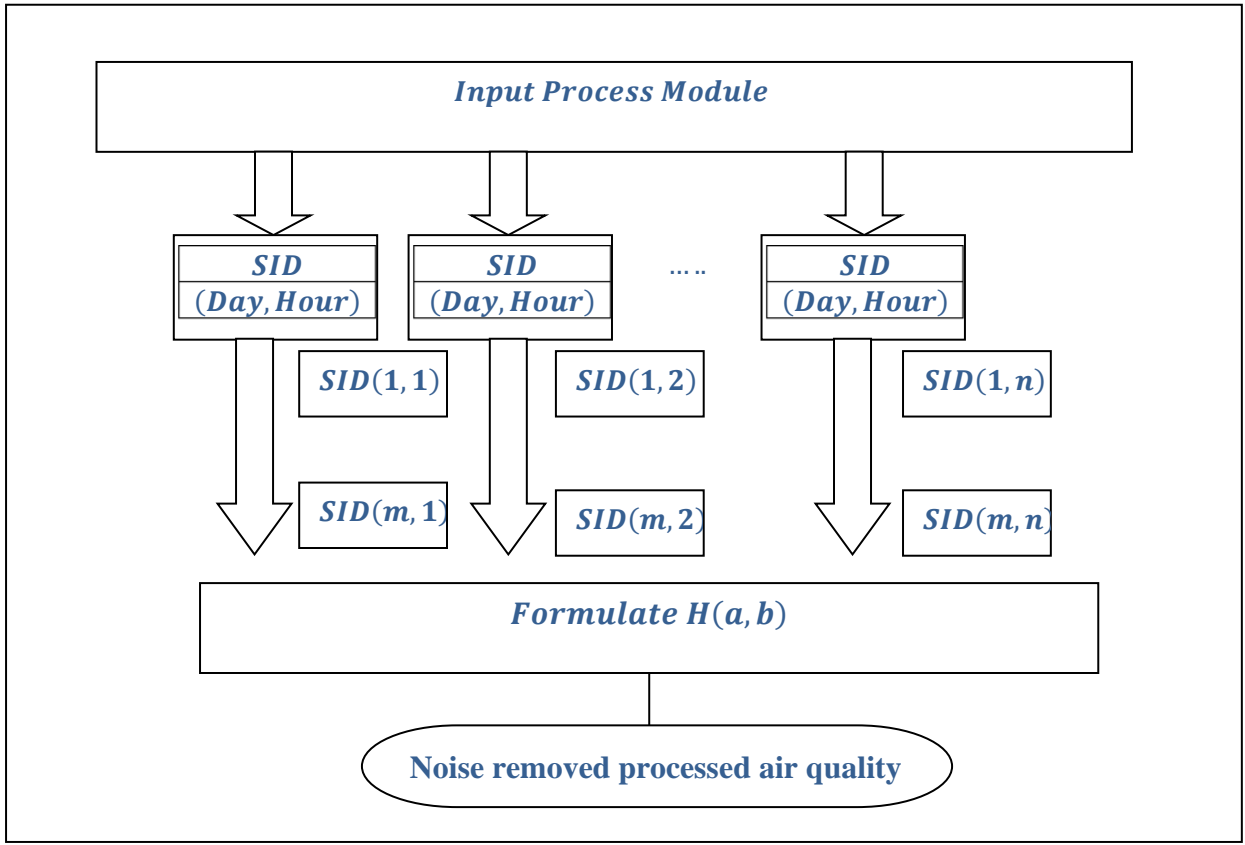

**Fig. 2 Structure of Discretized Hartley Transformation-based Preprocessing model**

### *3.2. Discretized Hartley Transformation-based Preprocessing model*

Preprocessing is very important to environmental data because it influences the results of any data investigation method and, therefore, the performance of any data mining algorithm towards problem investigation and parameter forecasting. Preprocessing is very important to environmental data because it influences the results of any data investigation method and, therefore, the performance of any data mining algorithm towards problem investigation and parameter forecasting. Preprocessing is essential as far as environmental data is concerned owing to the reason that it has a great impact on the data investigation method and, therefore, the performance of any machine learning technique towards air pollution forecasting. It is frequent in air quality data to come across noise and extract real trends and patterns. With the objective of obtaining comparable results in the proposed method, the rule for replacing using Discretized Hartley Transformation-based Preprocessing model is employed in our work. The Discretized Hartley Transformation-based Preprocessing model proposed in our work transforms real inputs into real outputs. Figure 2, given below, shows the structure of Discretized Hartley Transformation-based Preprocessing model.

As given in the above figure, to extract real trends and patterns and eliminate noise, various cities in India are obtained in the first hidden layer with the station details obtained from the corresponding ID. Then, outliers or noise are removed by means of transforming real inputs to real outputs via the Discretized Hartley Transformation function.

Let us consider the input air quality data '$D = D_1, D_2, \ldots, D_n$' obtained from sensors '$S = S_1, S_2, \ldots, S_n$' be sent into the first hidden layer. The input air quality data here is stored in the form of vector-matrix separately for the corresponding station according to '$Day$' and hour '$Hour$' as given below.

$$SID[Day] = Day\,[D_1, D_2, \ldots, D_n] \quad (1)$$

$$SID[Hour] = Hour\,[D_1, D_2, \ldots, D_n] \quad (2)$$

Then, from the above formulations from equations (1) and (2), two dimensional DHT pair is formulated as given below.

$$H(a,b) = \sum_{Day=1}^{m} \sum_{Hour=1}^{n} SID\,(Day, Hour)\,cas\left[2\pi\left(\frac{aDay}{P} + \frac{bHour}{Q}\right)\right] \quad (3)$$

$$SID\,(Day, Hour) = \sum_{Day=1}^{m} \sum_{Hour=1}^{n} H(a,b)\,cas\left[2\pi\left(\frac{aDay}{P} + \frac{bHour}{Q}\right)\right] \quad (4)$$

From the above equations (3) and (4), '$SID\,(Day, Hour)$' denotes the DHT pair averaged over the hours of the day, and the DHT pair averaged and '$H(a,b)$' represents the Hartley spectrum coefficient, Also '$P$' and '$Q$' represents the rows and columns of the air quality data and '$cas\,(\theta) = \cos(\theta) + \sin(\theta)$' and is mathematically formulated as given below.

$$PD = H(a,b) = \sum_{Day=1}^{m} \sum_{Hour=1}^{n} DIS(Day, Hour)\left(Cos\left[2\pi\left(\frac{aDay}{P} + \frac{bHour}{Q}\right)\right] + Sin\left[2\pi\left(\frac{aDay}{P} + \frac{bHour}{Q}\right)\right]\right) \quad (5)$$

From the results obtained in the above equation (5), the real inputs are transformed into real outputs, therefore eliminating the noise from air quality data. The pseudo-code representation of Discretized Hartley Transformation-based Preprocessing is given below.

**Algorithm 1. Discretized Hartley Transformation-based Preprocessing**

| |
|---|
| **Input**: Dataset '$DS$', IoT Devices or Sensors '$S = S_1, S_2, \ldots, S_n$', Features '$F = F_1, F_2, \ldots, F_n$', Air Quality data '$D = D_1, D_2, \ldots, D_n$' |
| **Output**: processed air quality data '$PD$' |
| Step 1: **Initialize** rows and columns of the air quality data '$P$' and '$Q$' <br> Step 2: **Begin** <br> Step 3: **For** each Dataset '$DS$' with Sensors '$S$' and air quality data '$D = D_1, D_2, \ldots, D_n$' <br> Step 4: Obtain vector matrix separately for the corresponding station based on the day '$Day$' and hour '$Hour$' as in equations (1) and (2) <br> Step 5: Formulate two dimensional HT pair as in equations (3) and (4) <br> Step 6: Obtain discretized HT pair as in equation (5) to transform real inputs to real outputs <br> Step 7: **Return** processed air quality data '$PD$' <br> Step 8: **End for** <br> Step 9: **End** |

As given in the above algorithm, with the objective of minimizing the error by eliminating the noise and extracting real trends and patterns, Discretized Hartley Transformation function is applied. With the raw air quality data obtained as input from the input layer, the first hidden layer performs the preprocessing by means of first separately forming the vector matrix. Next, the two-dimensional HT pair is structured based on day-wise and hourly air quality data. Finally, the transformation is performed by discretizing with the sine and cosine functions for the two-dimensional HT pair, therefore obtaining the processed air quality data.

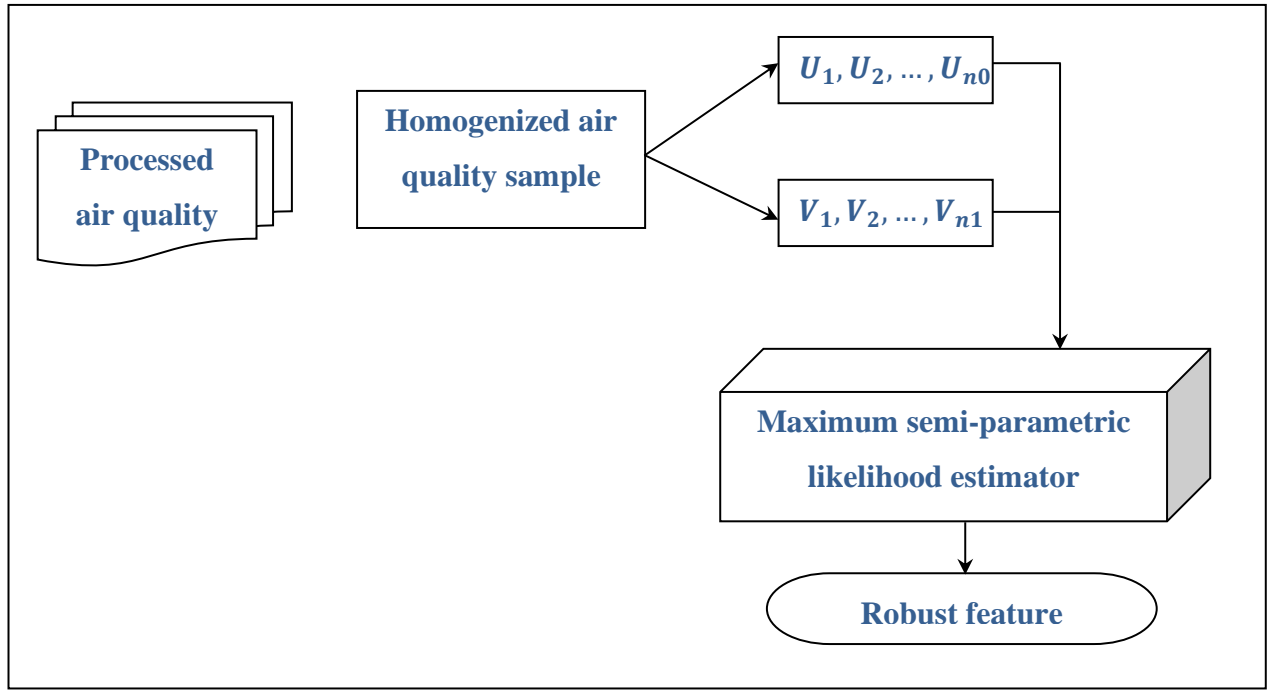

Fig. 3 Structure of Constrained Maximum Likelihood Linear Regression-based Feature Selection model

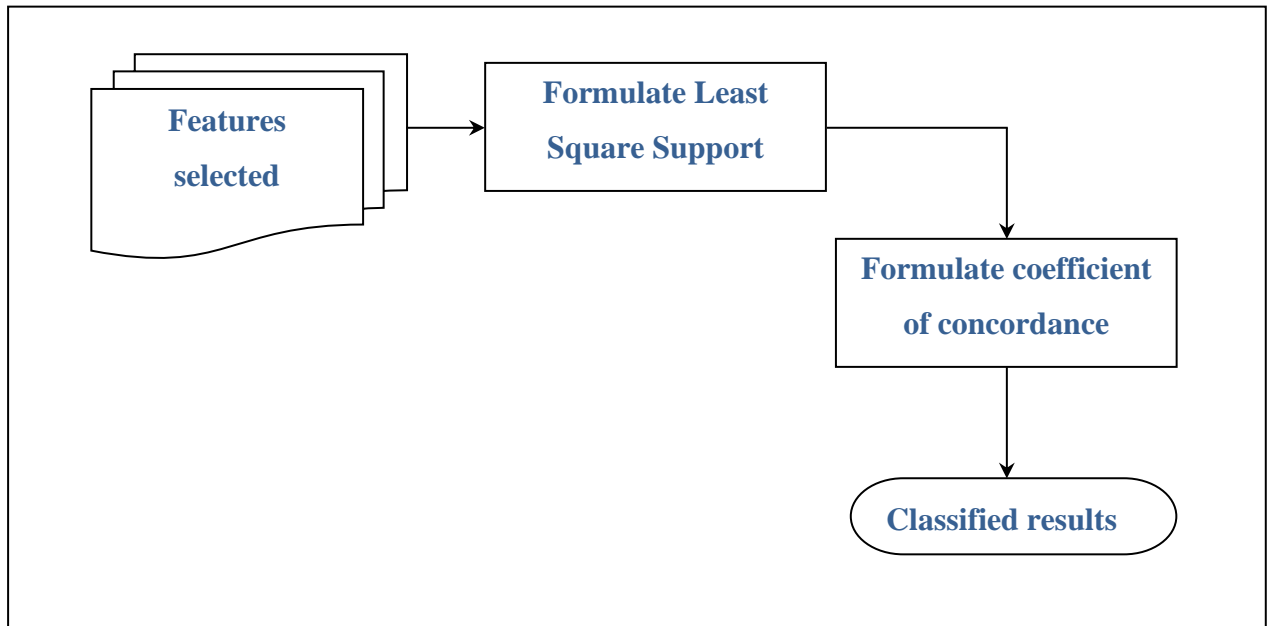

Fig. 4 Structure of Concordance Correlative Least Square Support Vector-based Classification

### 3.3. Constrained Maximum Likelihood Linear Regression-based Feature Selection model

For designing an efficient mechanism for tackling air pollution, it is essential to concentrate the endeavors on the pollutants most accountable for air pollution. Large numbers of processed air quality data features make the analysis a complicated task. Selection of the most essential and relevant features is the predominant step in air quality pollution monitoring and the decision-making process.

Several feature selection methods are present that aspire to minimize the number of dimensions in the processed air quality data to make the analysis simpler and more efficient. In this work, the feature selection process is carried out by employing Constrained Maximum Likelihood Linear Regression in the second hidden layer. Constrained Maximum Likelihood Linear Regression is one of the most popular models used for the selection of features by taking into consideration the correlative feature significance. The feature significance is utilized in describing which features are most relevant or significant for air quality index (AQI) prediction by means of the maximum likelihood function. Figure 3 below shows the structure of the Constrained Maximum Likelihood Linear Regression-based Feature Selection model.

The above figure shows that the processed air quality data comprises redundant information. The constrained Maximum Likelihood Linear Regression criterion is utilized to select the maximum likelihood regressors that carry the maximal non-redundant information for air quality pollution and monitoring[10]. Initially, the processed air quality data provided as input to the second hidden layer is split into two homogenized vectors, one for station-wise air quality data on an hourly basis and another for station-wise quality data on a daily basis. Next, the two homogenized vectors are subjected to a Maximum semi-parametric likelihood estimator to select relevant and robust features.

Let '$(U_1, U_2, ..., U_{n0})$' denote the arbitrary air quality sample data from '$Prob(u|V = 0)$' and '$(V_1, V_2, ..., V_{n1})$' denote the arbitrary air quality sample data from '$Prob(u|V = 1)$' respectively. Also, let '$(T_1, T_2, ..., T_n)$' represents the homogenized air quality sample data '$(U_1, U_2, ..., U_{n0}; V_1, V_2, ..., V_{n1})$'. Moreover, let '$(\alpha_0, \beta_0)$' denote the initial value of '$(\alpha, \beta)$'. Then, the logistic regression function for the homogenized air quality sample data with respect to the constrained '$(\alpha, \beta)$' is mathematically stated as given below.

$$Prob(V = 1|U = u) = \frac{\exp(\alpha + \beta^\tau T)}{1 + \exp(\alpha + \beta^\tau T)} \quad (6)$$

From the above equation (6), '$\alpha$' and '$\beta$', represents the scale values. Then, the maximum semi-parametric likelihood estimator of '$(\alpha_0, \beta_0)$' is mathematically formulated as given below.

$$\frac{\partial l(\alpha,\beta)}{\partial \alpha} = \sum_{i=1}^{n} \frac{\exp(U_1, U_2, ..., U_{n0})(\alpha + \beta^\tau T_i)}{\exp(\alpha + \beta^\tau T_i)} \quad (7)$$

$$\frac{\partial l(\alpha,\beta)}{\partial \beta} = \sum_{i=1}^{n} \frac{\exp(V_1, V_2, ..., V_{n1})(\alpha + \beta^\tau T_i)}{\exp(\alpha + \beta^\tau T_i)} \quad (8)$$

From the results of the above two equations (7) and (8), with a maximum semi-parametric likelihood estimator applied to the regression results, redundant features are eliminated, and relevant features are selected for further processing.

$$FS = \frac{\partial l(\alpha,\beta)}{\partial \alpha} \cup \frac{\partial l(\alpha,\beta)}{\partial \beta} \quad (9)$$

From the above equation (9), finally, the significant features selected, '$FS$', are obtained by combining the maximum semi-parametric likelihood estimator with respect to two constraints '$\partial \alpha$' and '$\partial \beta$', respectively. The pseudo-code representation of Constrained Maximum Likelihood Linear Regression-based Feature Selection is given below.

**Algorithm 2. Constrained Maximum Likelihood Linear Regression-based Feature Selection**

| |
|---|
| **Input**: Dataset '$DS$', IoT Devices or Sensors '$S = S_1, S_2, ..., S_n$', Features '$F = F_1, F_2, ..., F_n$', Air Quality data '$D = D_1, D_2, ..., D_n$' |
| **Output**: Computationally efficient feature selection '$FS$' |
| Step 1: **Initialize** processed air quality data '$PD$' |
| Step 2: **Initialize** arbitrary air quality sample data '$(U_1, U_2, ..., U_{n0})$' (i.e., processed station data at hourly basis |
| Step 3: **Initialize** arbitrary air quality sample data '$(V_1, V_2, ..., V_{n0})$' (i.e., processed station data at daily basis) |
| Step 4: **Begin** |
| Step 5: **For** each Dataset '$DS$' with Sensors '$S$' and processed air quality data '$PD$' |
| Step 6: Evaluate the logistic regression function for the homogenized air quality sample data as in equation (6) |
| Step 7: Evaluate the maximum semi-parametric likelihood estimator as in equations (7) and (8) |
| Step 8: **Return** features selected '$FS$' |
| Step 9: **End for** |
| Step 10: **End** |

As given in the above algorithm, arbitrarily processed air quality is obtained to obtain computationally efficient features. Second, the logistic regression function is applied separately to make the analysis easier and simpler. Finally, the maximum semi-parametric likelihood estimator is modeled to select the significant and computationally efficient feature. [PM10, CO, O3]

*3.4. Concordance Correlative Least Square Support Vector-based Classification model*

Finally, with the selected significant features, classification has to be performed for accurate air pollution forecasting with the objective of minimizing the false positive rate. With the aid of the Air Quality Index (AQI), pollutant levels in the air can be determined. The range of AQI varies between '$0\ and\ 500$', with a higher number referring to lower air quality.

Numerous classification methods have been designed for air pollution forecasting; however, with AQI limits varying in nature, the inappropriate classification may cause adverse effects on human health. Hence, a model is required for precise air pollution forecasting with appropriate classification, therefore resulting in the improvement of the false positive rate. In this work, the Concordance Correlative Least Square Support Vector-based Classification model is applied in the third hidden layer to measure the nonlinear relationship between input variables and output variables with the least error.

With the measurement of the nonlinear relationship between input variables and output variables, the false positive rate or assessing the exposure of the target population to specific air pollutants analyzed is said to be precise. Moreover, applying the Concordance Correlative function establishes a nonlinear relationship between input variables and output variables, therefore obtaining the target population for specific air pollutants precisely. Figure 4 shows the structure of the Concordance Correlative Least Square Support Vector-based Classification model.

As shown in the figure 4, the Concordance Correlative Least Square Support Vector-based Classification model provides a linear equation solution with an enhancement in the objective function of conventional SVM. We employ '$FS_k$' as the three features selected with maximum semi-parametric likelihood estimator. Next, the selected features are subjected to the least square support vector. With the resultant values, the coefficient of concordance is applied to obtain the classified results. Let us consider the Least Square Support Vector-based Classification written as given below.

$$y(FS) = \omega^T \varphi(FS) + B \quad (10)$$

From the above equation (10), '$\varphi(FS)$' forms the nonlinear mapping function for the selected features, with '$\omega$' forming the weight and '$B$' the bias, respectively. The equation is then subjected to minimum correlation with respect to error '$Err$' as given below.

$$\min(\omega, Err) = \tfrac{1}{2}\omega^T\omega + \tfrac{1}{2}\gamma \sum_{k=1}^{n} Err_k^2 \quad (11)$$

$$Subject\ to\ y_k = \omega^T \varphi(FS_k) + B + Err_k \quad (12)$$

From the above equations (11) and (12), '$\gamma$' forms the regularization parameter (i.e., setting index value) and '$Err$' represents the error, respectively. The optimization equation for the coefficient of concordance using Kendall's Rank Correlation Coefficient is mathematically formulated as given below.

$$AQI = \tau = \frac{(N_{CP})\varphi(FS_k) - (N_{DP})\varphi(FS_k)}{\frac{n(n-1)}{2}} \quad (13)$$

From the above equation (13), any pair of observations '$(Tr_i, Ts_i)$' and '$(Tr_j, Ts_j)$' are said to be concordant if the sort order of '$(Tr_i, Tr_j)$' and '$(Ts_i, Ts_j)$' agrees, otherwise they are said to be discordant, '$\frac{n(n-1)}{2}$' denotes the binomial coefficient to select two features from '$n$' features. The denominator here represents the total number of pair combinations, hence coefficient ranging in between '$-1 \leq \tau \leq 1$'. The pseudo code representation for Concordance Correlative Least Square Support Vector-based Classification is given below.

| Algorithm 3 Concordance Correlative Least Square Support Vector-based Classification |
|---|
| Input: Dataset '$DS$', IoT Devices or Sensors '$S = S_1, S_2, \ldots, S_n$', Features '$F = F_1, F_2, \ldots, F_n$', Air Quality data '$D = D_1, D_2, \ldots, D_n$' |
| Output: |
| Step 1: **Initialize** processed air quality data '$PD$', features selected '$FS$' |
| Step 2: **Begin** |
| Step 3: **For** each Dataset '$DS$' with Sensors '$S$', processed air quality data '$PD$' and features selected '$FS$' |
| Step 4: Formulate Least Square Support Vector-based Classification as in equation (10) |
| Step 5: Evaluate minimum correlation with respect to error as in equations (11) and (12) |
| Step 6: Measure Kendalls Rank Correlation Coefficient as in equation (13) |
| Step 7: **If** '$\tau\ lies\ between -1\ and -0.5$' |
| Step 8: **Then** air quality is very poor |
| Step 9: **End if** |
| Step 10: **If** '$\tau\ lies\ between -5\ and\ 0$' |
| Step 11: **Then** air quality is poor |
| Step 12: **End if** |
| Step 13: **If** '$\tau\ lies\ equals\ 0$' |
| Step 14: **Then** air quality is good |
| Step 15: **End if** |
| Step 16: **If** '$\tau\ lies\ between\ 0\ and +0.5$' |
| Step 17: **Then** air quality is satisfactory |
| Step 18: **End if** |
| Step 19: **If** '$\tau\ lies\ between +0.5\ and\ 1$' |
| Step 20: **Then** air quality is moderate |
| Step 21: **End if** |
| Step 22: **Else** |
| Step 23: Air quality is very severe |
| Step 24: **End if** |
| Step 25: **End for** |
| Step 26: **End** |

As given in the above algorithm, to improve the false positive rate or analyze the air quality as it is, not only is high correlation required but also concordance with each other.

Therefore, first, the Least Square Support Vector for the respective selected features is made. Second, to satisfy true positivity, minimum correlation with respect to error should be maintained. Hence, the correlative concordance function employing Kendall's Rank Correlation Coefficient is formulated. With this, air pollution forecasting is done with high accuracy.

## 4. Experimental setup

Discretized Regression and Least Square Support Vector (DR-LSSV) based air pollution monitoring and control for IoT networks are used to evaluate the proposed method. Several experiments are conducted with Deep AIR [1] and Variational Auto Encoder (VAE) based on the innovative Integrated Multiple Direct Attention Deep Learning architecture (IMDA) (VAE-IMDA) [2]. We also compared the performance of the proposed DR-LSSV method with [1] and [2] in JAVA language using an air quality dataset to measure the efficiency in terms of air pollution forecasting accuracy, air pollution forecasting time and false positive rate.

## 5. Discussion

This section provides the results analysis of three distinct parameters, air pollution forecasting accuracy, air pollution forecasting time and false positive rate.

### 5.1. Results Analysis of Air Pollution Forecasting Accuracy

The first set of experiments is conducted to analyze the performance of the proposed method, DR-LSSV, with respect to air pollution forecasting accuracy in forecasting air pollution. With Air Quality Index being applied as the measure by the Indian government to quantify air pollution, air pollution forecasting demands sophisticated monitoring tools and mechanisms along with advanced models to estimate time-related pollutant data. Hence, accurate air quality forecasting is considered to be pivotal for systematic pollution control. The air pollution forecasting accuracy is measured as given below.

$$APF_{acc} = \sum_{i=1}^{n} \frac{D_{AF}}{D_i} \quad (14)$$

From the above equation (14), the air pollution forecasting accuracy. '$APF_{acc}$' is evaluated based on the number of air quality data involved in a simulation activity. '$D_i$' and the data being accurately forecasted. '$D_{AF}$'. It is measured in terms of percentage (%). Table 2 shows the air pollution forecasting accuracy performance comparison of the three methods, DR-LSSV, Deep-AIR [1], and VAE-IMDA [2], based on AQI measurements.

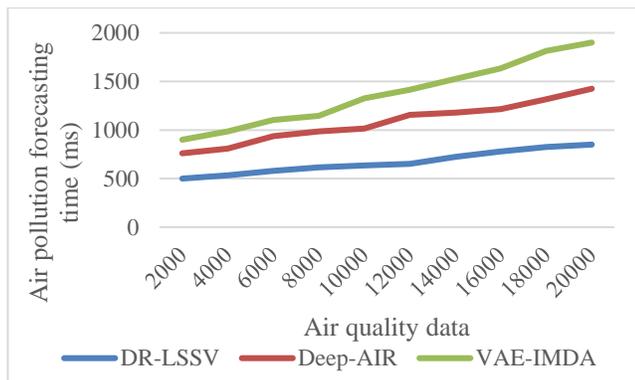

**Fig. 5 Performance comparison of air pollution forecasting accuracy**

**Table 1. Air pollution forecasting accuracy performance comparison of the proposed DR-LSSV method using air quality dataset from India**

| Air quality data | Air pollution forecasting accuracy (%) | | |
|---|---|---|---|
| | DR-LSSV | Deep-AIR | VAE-IMDA |
| 2000 | 98.25 | 96.5 | 95.25 |
| 4000 | 98.05 | 96 | 94.25 |
| 6000 | 98 | 95.75 | 94.15 |
| 8000 | 97.55 | 92.15 | 89.15 |
| 10000 | 97.15 | 91 | 87.35 |
| 12000 | 97 | 90 | 85.55 |
| 14000 | 96.85 | 85 | 82.15 |
| 16000 | 96.25 | 83.15 | 78.55 |
| 18000 | 96 | 82 | 76.35 |
| 20000 | 95.35 | 80 | 72 |

Figure 5 shows the air pollution forecasting accuracy measured using the three methods DR-LSSV, Deep-AIR [1] and VAE-IMDA [2] considering the AQI measurements. Sample results of ground truth and predicted air pollution forecasting accuracy for the case of air quality data in India are shown in Figure 5. The proposed DR-LSSV method provides air quality data accuracy forecasting via smooth labeling, and it was able to forecast multiple classes based on Kendall's Rank Correlation Coefficient. The differences between any pair of observations in the output forecasting and ground-truth forecasting for each distinct air quality data ranging between 2000 and 20000 are also provided. On average minimum accuracy differs from 95.35% to 98.25% using DR-LSSV, 80% to 96.5% using [1] and 72% to 95.25% using [2], respectively. With this range, the DR-LSSV method was found to be improved in ensuring accuracy level by 9% compared to [1] and 14% compared to [2]. The reason behind the improvement using the DR-LSSV method was selecting the computationally efficient features using the logistic regression function and maximum semi-parametric likelihood estimator. As a result, with accurate air pollution forecasting, preventive measures in various cities can be taken to control the hazardous factor, resulting in a healthy environment.

### 5.2. Results Analysis of Air Pollution Forecasting Time

By analyzing historical data sets like air quality data in India, correlations can be made between pollution levels and meteorological data variables, i.e., air quality index. Pollution levels can be measured with the resulting values, ensuring air pollution forecasting efficiently. During this process, a small portion of time is utilized, and that time is referred to as the air pollution forecasting time. The air pollution forecasting time is mathematically formulated as given below.

$$APF_{time} = \sum_{i=1}^{n} D_i * Time\ [\tau] \quad (15)$$

From the above equation (15), the air pollution forecasting time. '$APF_{time}$' is measured based on the number of air quality data in concern. '$D_i$' and the time involved in the forecasting process '$Time\ [\tau]$'. It is measured in terms of milliseconds (ms). Table 3 illustrates the air pollution forecasting time comparative results of the considered methods, DR-LSSV, Deep-AIR [1] and VAE-IMDA [2], respectively.

**Table 2. Air pollution forecasting time performance comparison of the proposed DR-LSSV method using air quality dataset from India**

| Air quality data | Air pollution forecasting time (ms) | | |
|---|---|---|---|
| | DR-LSSV | Deep-AIR | VAE-IMDA |
| 2000 | 500 | 760 | 900 |
| 4000 | 535 | 810 | 985 |
| 6000 | 580 | 935 | 1105 |
| 8000 | 615 | 985 | 1145 |
| 10000 | 635 | 1015 | 1325 |
| 12000 | 650 | 1155 | 1415 |
| 14000 | 725 | 1180 | 1525 |
| 16000 | 780 | 1215 | 1635 |
| 18000 | 825 | 1315 | 1815 |
| 20000 | 850 | 1425 | 1900 |

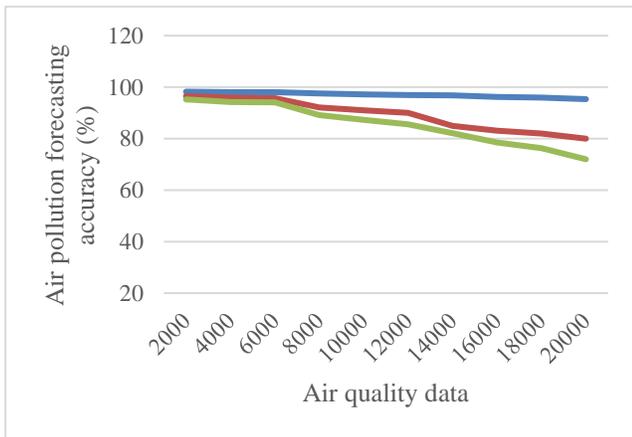

**Fig. 6 Performance comparison of air pollution forecasting time**

To study the influence of air pollution forecasting time on distinct numbers of air quality data ranging between 2000 and 20000, experiments were performed with different thresholds (i.e., stations being observed both on a daily and hourly basis). The trends of the forecasting time involved on both a daily and hourly basis can be seen in Figure 6. We note that as the number of air quality data is increased, the percentage of maximum semi-parametric likelihood estimator decreases, and the percentage of air pollution forecasting time decreases steadily. However, the homogenized air quality sample in terms of air quality data increases with the increase in the improperly processed results. This, in turn, results in the minimization of air pollution forecasting time using the DR-LSSV method by 38% compared to [1] and 50% compared to [2], respectively.

### 5.3. Results Analysis of the False Positive Rate

Finally, the third-second set of experiments aims to analyze the performance of DR-LSSV concerning the false positive rate involved in air pollution monitoring. Finally, the false positive rate also referred to as the false alarm ratio, is measured. The false positive rate in air pollution forecasting refers to the probability of falsely rejecting the null hypothesis (i.e., falsely rejecting air pollutants in consideration for measuring the air quality index) for a particular test (i.e., air pollution forecasting). In other words, the false positive rate is measured as the percentage ratio between the number of negative events wrongly categorized as positive (i.e., false air pollutants wrongly involved in the measurement of air quality index) and the total number of air quality data.

$$FPR = \frac{FP}{(FP+TN)} \qquad (16)$$

From the above equation (16), the false positive rate '$FPR$' is measured based on the false positive (i.e., wrong inclusion of air pollutants) '$FP$' and the true negatives (i.e., actual wrong air pollutants) '$T$'. Finally, table 4, given below, lists the false positive rate.

**Table 3. False positive rate performance comparison of the proposed DR-LSSV method using air quality dataset from India**

| Air quality data | False positive rate (%) | | |
|---|---|---|---|
| | DR-LSSV | Deep-AIR | VAE-IMDA |
| 2000 | 0.375 | 0.45 | 0.675 |
| 4000 | 0.395 | 0.485 | 0.735 |
| 6000 | 0.41 | 0.525 | 0.785 |
| 8000 | 0.435 | 0.545 | 0.815 |
| 10000 | 0.485 | 0.595 | 0.83 |
| 12000 | 0.515 | 0.605 | 0.845 |
| 14000 | 0.535 | 0.625 | 0.855 |
| 16000 | 0.57 | 0.645 | 0.875 |
| 18000 | 0.585 | 0.715 | 0.915 |
| 20000 | 0.615 | 0.735 | 0.925 |

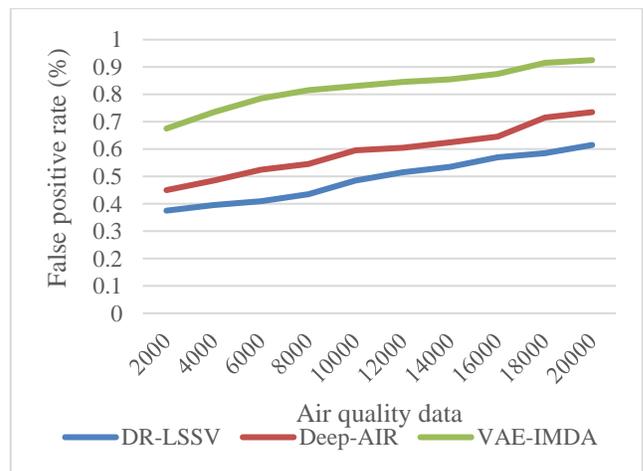

**Fig. 7 Performance comparison of false positive rate**

Finally, figure 7 given above illustrates the false positive rate in the y-axis for distinct numbers of air quality data ranging between 2000 and 20000 in the x-axis, respectively. From the above figure, increasing the number of air quality data in the above simulation process results in an increase in the false positive rate using the three methods, DR-LSSV, Deep-AIR [1] and VAE-IMDA [2], respectively. However, with simulations performed using 2000 numbers of air quality data, 15 numbers of air quality data were wrongly assessed using DR-LSSV, 22 numbers air quality data were wrongly assessed using [1], and 27 numbers of air quality data wrongly assessed using [2], the overall false positive rate using the three methods were found to be 0.375, 0.45 and 0.675 respectively. With this result, the false positive rate using the DR-LSSV method was found to be comparatively lesser than [1] and [2]. The reason behind the improvement was the application of the Concordance Correlative Least Square Support Vector-based Classification algorithm in the third hidden layer. Here, multiple classes were evolved according to the resultant values in the AQI based on the processed feature selected. As a result, the falsification of air quality data in the DR-LSSV method was found to be comparatively lesser by 17% compared to [1] and 41% compared to [2].

## 6. Conclusion

Existing methods to forecast air pollution method to detect the level of pollution in the atmosphere using attention mechanism and spatio-temporal features employ 1 to 1 convolution layers from the nonlinear approximations attributes. Our proposed method is capable of obtaining computationally efficient and robust features due to the discretizing of air quality data using sine and cosine functions on both a daily and hourly basis. It automatically extracts strong features from the Constrained Maximum Likelihood function. The Concordance Correlative Least Square Support Vector-based Classification model is applied with the aid of selected features to reduce the false positive rate or falsely select the air pollutants for classification. Finally, multiclass classification results were obtained by means of Kendall's Rank Correlation Coefficient. With this, the DR-LSSV method precisely and accurately forecasts air pollution and minimizes the falsification of air pollutants for further analysis in a timely manner. Experiments on real air quality data in the India dataset show that the DR-LSSV method is superior to the existing method regarding air pollution forecasting accuracy, air pollution forecasting time and false positive rate.